\documentclass[manuscript,review=false]{acmart}

\AtBeginDocument{%
  \providecommand\BibTeX{{\normalfont B\kern-0.5em{\scshape i\kern-0.25em b}\kern-0.8em\TeX}}}
\setcopyright{none}
\renewcommand\footnotetextcopyrightpermission[1]{}
\acmConference[CHI '27]{CHI Conference on Human Factors in Computing Systems}{2027}{TBD}
\acmBooktitle{CHI Conference on Human Factors in Computing Systems}

\usepackage{booktabs}
\usepackage{xcolor}
\usepackage{tabularx}
\newcolumntype{L}{>{\raggedright\arraybackslash}X}
\usepackage{pifont}
\usepackage{hyperref}
\usepackage{microtype}
\usepackage{comment}
\newcommand{\cmark}{\textcolor{green!55!black}{\ding{51}}}
\newcommand{\xmark}{\textcolor{red!75!black}{\ding{55}}}

\begin{document}

\title{Toward Human-in-the-Loop Robot Failure Recovery: Bridging Communication Gaps in Human-Robot Collaboration}

\author{Promise Ekpo}
\affiliation{%
  \institution{Cornell University}
  \city{New York}
  \state{New York}
  \country{USA}}

\author{Teju Vijay}
\authornote{Teju Vijay and Dhruv Mandalik contributed equally to this work.}
\affiliation{%
  \institution{Cornell University}
  \city{New York}
  \state{New York}
  \country{USA}}

\author{Dhruv Mandalik}
\authornotemark[1]
\affiliation{%
  \institution{Cornell University}
  \city{New York}
  \state{New York}
  \country{USA}}

\author{Tisha Jain}
\affiliation{%
  \institution{Cornell University}
  \city{New York}
  \state{New York}
  \country{USA}}

\author{Arman Ibrayeva}
\affiliation{%
  \institution{Cornell Tech}
  \city{New York}
  \state{New York}
  \country{USA}}

\author{Sunishka Sil}
\affiliation{%
  \institution{ Cornell Tech}
  \city{New York}
  \state{New York}
  \country{USA}}

\author{Stefanie A. Tellex}
\affiliation{%
  \institution{Brown University}
  \city{Providence}
  \state{Rhode Island}
  \country{USA}}

\author{Angelique Taylor}
\affiliation{%
  \institution{Cornell Tech}
  \city{New York}
  \state{New York}
  \country{USA}}

\renewcommand{\shortauthors}{Ekpo et al.}

\begin{abstract}
Robots can recover from failures by asking bystanders for help, but effective human-in-the-loop recovery requires communication that accounts for differences in people's knowledge. Prior inverse-semantics work generates requests using a single listener model, leaving differences in listener knowledge untested. We introduce Listener Differences in Human-Robot Interaction (LD-HRI), a game, dataset, and benchmark that evaluates speakers through human listener performance. Our evaluation examines request properties, large language model (LLM) speakers, and inverse-semantics request-selection algorithms under controlled differences in listener information. The corpus contains 446 human-written requests and 1{,}302 listener trials. We additionally evaluated 24 frozen LLM-written requests with 70 human listeners across 560 trials. Novice success is descriptively higher with model-written requests across all four tasks, yet both request sources leave substantial expert--novice gaps, including 16 percentage points for LLM requests. LD-HRI makes these gaps measurable, providing a foundation for designing more robust communication in human-robot and human-agent interaction.
\end{abstract}



\begin{teaserfigure}
  \centering  
  \includegraphics[width=0.9\textwidth]{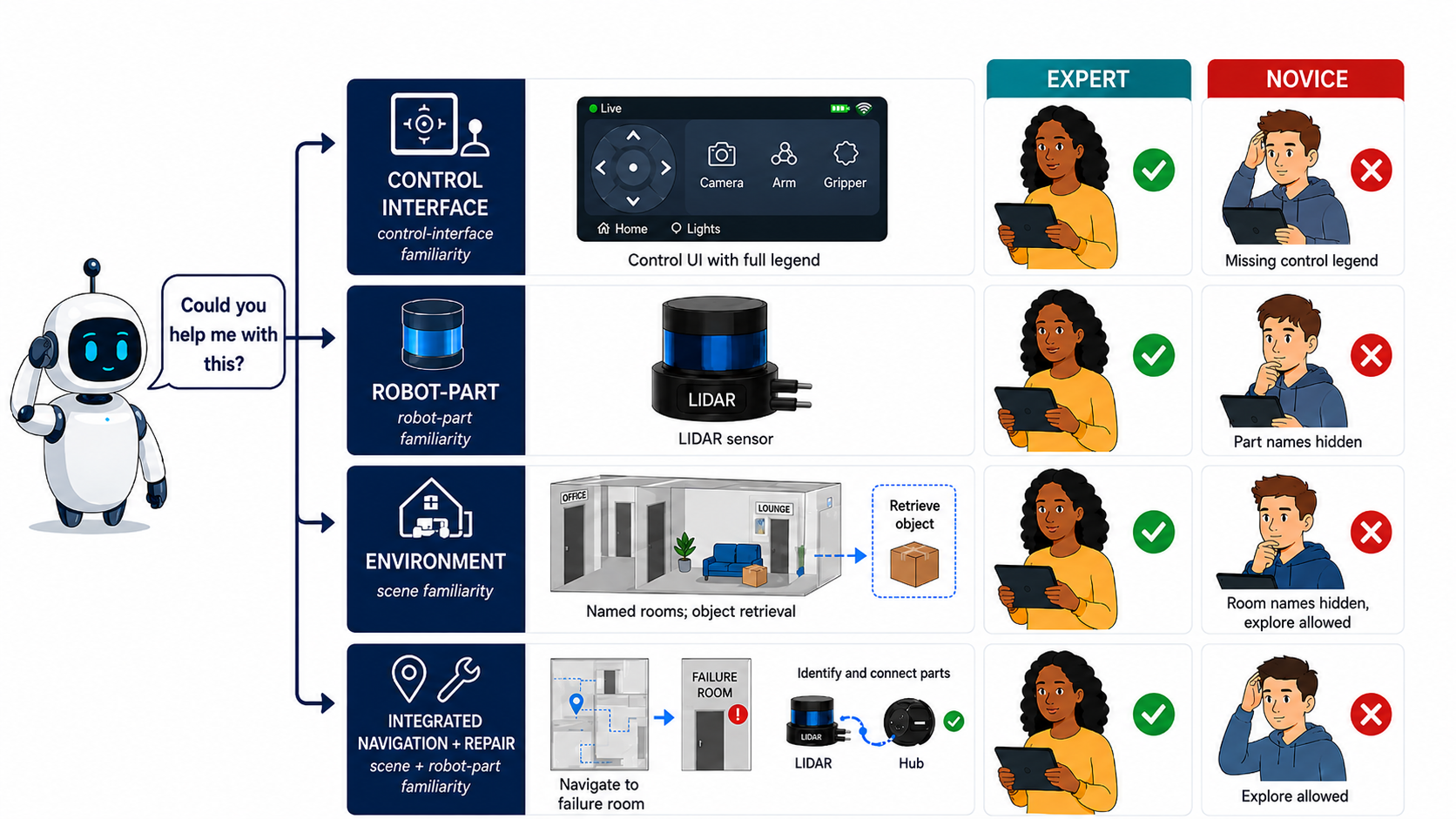}
  \caption{\textbf{Robot help requests can lead to different outcomes depending on listener's knowledge. Listener Differences in Human-Robot Interaction (LD-HRI) benchmark creates controlled differences in Listener knowledge (information). Teleoperation task manipulates robot control interface information, Hardware Repair manipulates robot parts information, Object Retrieval manipulates scene information, and Integrated Navigation and Repair combines scene and robot-parts information. The same Speaker request is evaluated with Expert and Novice Listeners.}}
  \Description{The same help request can lead to different outcomes depending on the listener's knowledge. LD-HRI varies three forms of Listener familiarity across four tasks, and the same Speaker request is evaluated with Expert and Novice Listeners.}
  \label{fig:teaser}
\end{teaserfigure}

\maketitle


\section{Introduction}

Robots are increasingly deployed alongside humans for everyday tasks such as cooking~\cite{noh_yummy_2024} and cleaning~\cite{mohan_design_2022}, as well as safety-critical applications such as search and rescue~\cite{chitikena_robotics_2023} and critical care~\cite{teng_use_2022}. While robots may operate autonomously, failures can occur that require assistance from a human bystander who may or may not possess relevant knowledge. For example, a malfunctioning robot part, such as a light detection and ranging (LiDAR) sensor, might need to be replaced because the robot cannot repair itself. In these cases, the robot (i.e speaker) must communicate effectively so that the human (i.e listener) understands enough to take the optimal action for failure recovery.

The difficulty is that there may be a mismatch between the robot's generated help request and the potential helper's understanding of the robot or environment. A request such as ``fetch the LiDAR sensor from the storage closet'' may be sufficient for someone who recognizes the robot part and knows its location. An unfamiliar helper may need a description of the sensor and directions to the storage closet. This example shows a perspective-taking problem in human-robot interaction (HRI). The effectiveness of a help request depends on the information provided to a listener.

Prior work on generating robot help requests during failure recovery models the effect of language on human actions~\cite{tellex_asking_2014} using a single listener model. Other work focuses on differences in the visual or spatial information available to a person~\cite{suhr_executing_2022,udagawa_natural_2019,achlioptas_referit3d_2020}. While these studies are useful for understanding differences in viewpoint in perspective-taking, potential human observers may differ in additional factors, such as familiarity with robots or the environment.
 Kojima et al.~\cite{kojima_continual_2021} study communication as followers gain experience, while Takmaz et al.~\cite{takmaz_speaking_2023} manipulate listener knowledge computationally. 
 These approaches leave a need for a human dataset on robot help requests under controlled differences in scene, robot-part, and control-interface information.
To address this gap, we introduce \textit{Listener Differences in Human-Robot Interaction} (LD-HRI), a web-based communication game, dataset, and benchmark for evaluating human speakers, large language model (LLM) speakers, and help-request selection algorithms through human listener performance. The game compares outcomes for the same request across expert and novice listener conditions. Across four tasks, LD-HRI manipulates control-interface, robot-part, and scene information, with the fourth task combining scene and robot-part information (Figure~\ref{fig:teaser}). Expert listeners receive task-specific reference information, while novice listeners complete the same task with that information withheld. These conditions represent differences in available visual information, not classifications of participants' prior expertise.

This design provides a controlled way to study perspective-taking in robot communication beyond differences in spatial viewpoint. The dataset contains 446 human-authored requests and 1{,}302 listener trials from 240 participants, including speakers and listeners. Expert success was 76.2\%, compared with 47.9\% for novices. The expected expert advantage provides evidence that the information manipulation matters for task performance. The same request is evaluated with expert and novice listeners, allowing us to compare the help it provides to people with different task information. Human speakers were informed about task information their listeners might lack, yet an expert--novice gap remained.

In Hardware Repair, requests that combined part names with visible descriptions (shape, color) were associated with higher novice success than requests using names alone. In Object Retrieval, listeners could discover some missing scene information through exploration. These findings motivate requests that provide information the listener cannot easily recover while completing the task. Novices also reported higher workload, and clearer requests were associated with lower workload in both listener conditions.

We additionally evaluate three LLMs as speakers using 24 frozen requests and 70 human listeners across 560 trials. Novice success is higher with model-written requests across all four tasks, but both human-written and model-written requests leave an expert--novice gap. LLM-written requests produced 73.9\% expert and 57.5\% novice success, a difference of 16.4\%. These separately collected studies do not establish a causal advantage for either speaker type. Together, the human and LLM results show that the evaluated requests do not fully bridge differences in listener information. Future methods can use LD-HRI to evaluate improvements through task success, workload, and interaction logs.

LD-HRI also evaluates request-selection methods using human listener outcomes. We compare adaptations of the inverse-semantics reference policies $S_1$ and $S_2$~\cite{tellex_asking_2014}. The $S_1$ policy uses task information expressed in the request, while $S_2$ additionally uses a general model of listener success. Reporting their outcomes separately for expert and novice listeners supports evaluation of algorithms across differences in listener information.

We organize the benchmark around four research questions (RQs).
\begin{itemize}
\item \textbf{RQ1.} How do expert and novice information conditions affect listeners' task performance and experience across tasks?
\item \textbf{RQ2.} Which properties of human-written requests are associated with listener success?
\item \textbf{RQ3.} How effective are LLM-generated help requests for human listeners compared with human-generated requests?
\item \textbf{RQ4.} How do requests selected by the inverse-semantics reference policies $S_1$ and $S_2$ perform for expert and novice human listeners?
\end{itemize}

Following the contribution types of Wobbrock and Kientz~\cite{wobbrock2016contributions}, we make four contributions.

\begin{enumerate}
\item \textbf{An artifact contribution.}
We provide the LD-HRI game and evaluation platform to support future research on listener-aware robot communication. The game controls the information available to expert and novice listeners and records task outcomes, actions, workload, and listener perceptions.

\item \textbf{A dataset contribution.}
We provide the LD-HRI corpus of 446 human-authored robot help requests and 1{,}302 listener trials, together with expert and novice outcomes where available, workload, perceptions of clarity and usefulness, and interaction logs. The corpus supports comparisons of the same request across listener information conditions.

\item \textbf{An empirical contribution.}
We show that both human-written and LLM-generated requests leave an expert--novice performance gap, and that requests describing visible component properties are associated with higher novice success in Hardware Repair. These findings suggest that robot help requests should explain unfamiliar terms and provide the information people need to act. Evaluations should measure both task success and workload for expert and novice listeners.

\item \textbf{A methodological contribution through a benchmark.}
We evaluate speakers through human listener performance across controlled differences in task information. Both human-written and LLM-written requests leave an expert--novice gap, providing a benchmark for developing help requests that work for people with different knowledge.
\end{enumerate}

\section{Related Work}

\subsection{Robot Help Requests and Listener Models}
Prior work has studied how robots can request human assistance when they cannot complete a task independently.
Tellex et al.~\cite{tellex_asking_2014} introduced inverse semantics, which selects the language of the help requests by maximizing the likelihood of the human taking the optimal robot recovery given the environment model.
The $S_0$ algorithm uses a fixed request and models neither the environment nor the listener. $S_1$ policy adds an environment model and selects language that matches the desired action in the current scene, but does not model how the listener will interpret it. $S_2$ policy models both the environment and the Listener's language understanding, allowing the robot to prefer more specific requests when shorter descriptions are ambiguous~\cite{tellex_asking_2014}.
Recent approaches have focused on when a robot should request assistance or clarification~\cite{caglar_help_nodate,ren_robots_2023}.
Wachowiak et al.~\cite{wachowiak_what_2026} further show that novice and experienced users ask household robots different questions, providing evidence that the level of user experience can shape robot communication.
Together, these studies motivate reasoning about the person involved in robot recovery, but they do not account for how human listeners with varying scene, or robot interface knowledge perform when the robot help-request is held constant.
HCI research has also examined how robots recruit help from people in public settings. Yu et al.'s DIS study~\cite{yu2024playful} compared verbal, emotional, and playful help-seeking strategies and found that the strategy shaped participants' willingness to help and their experience of the robot. LD-HRI complements this work by focusing on the content of a request and whether a person can act on it with the information available to them.

\subsection{Perspective Taking and Referring Expressions}
Prior grounded communication datasets manipulate visual or role-based information asymmetries. CerealBar contrasts overhead and first-person views~\cite{suhr_executing_2022}, while OneCommon provides partially overlapping visual information~\cite{udagawa_natural_2019}. GuessWhat?! and TEACh assign hidden target or task information across roles~\cite{vries_guesswhat_2017,padmakumar_teach_2021}. RefCOCO and ReferIt3D study referring expressions under visual and spatial ambiguity~\cite{kazemzadeh_referitgame_2014,mao_generation_2016,achlioptas_referit3d_2020}, and Kojima et al.~\cite{kojima_continual_2021} examine communication as followers gain experience. As summarized in Table~\ref{tab:related-datasets}, these settings establish that perspective and experience shape communication, but they do not provide paired human outcomes for the same request under controlled differences in scene, robot-part, and control-interface information.

\subsection{Listener Aware Language Generation}
Listener aware language generation asks whether a Speaker should adapt an utterance to what a Listener is expected to know.
Rational Speech Acts produces pragmatic language  through reasoning about a listener~\cite{frank_predicting_2012}.
Takmaz et al.~\cite{takmaz_speaking_2023} model a knowledgeable Speaker communicating with listeners whose training exposure differs across domains and show that modeling Listener knowledge can improve communicative success. However, their Listener differences are imposed computationally across model variants instead of being experimentally assigned to human participants.
Existing instruction following benchmarks such as ALFRED~\cite{shridhar_alfred_2020} and DialFRED~\cite{gao_dialfred_2022} also evaluate grounded task completion but do not compare paired human outcomes for the same request under different listener information conditions.
Thus, prior work establishes that perspective, experience, and listener knowledge matter for communication, but does not provide paired human evidence needed to investigate the reason a robot help request works for one information condition and fails for another.

\begin{table*}[t]
  \centering
  \caption{\textbf{Comparison of related human--human communication
    datasets against the requirements of listener-aware help-request
    generation.}}
  \Description{Comparison table of ten prior datasets against our corpus
    across four properties: asymmetry source, listener familiarity as a
    controlled independent variable, paired novice and expert outcomes on
    the same utterance, behavioral outcomes, and workload or subjective
    ratings.}
  \label{tab:related-datasets}
  \small
  \resizebox{\textwidth}{!}{%
  \begin{tabular}{@{}l l c c l l@{}}
    \toprule
    \textbf{Dataset} & \textbf{Asymmetry source}
      & \textbf{\shortstack{Familiarity\\as controlled IV}}
      & \textbf{\shortstack{Paired outcomes,\\\emph{same} utterance}}
      & \textbf{\shortstack{Behavioral\\outcomes}}
      & \textbf{\shortstack{Workload +\\subjective ratings}} \\
    \midrule
    Cards~\cite{potts_goal-driven_2012}            & Roles (advisor/advisee)           & \xmark\ (emergent)             & \xmark & Partial (dialogue-focused)   & \xmark \\
    CerealBar~\cite{suhr_executing_2022}    & Visual (overhead vs.\ ground) & \xmark\ (qualification filter) & \xmark & \cmark                       & \xmark \\
    OneCommon~\cite{udagawa_natural_2019}    & Visual (private dot subset)   & \xmark & \xmark & Partial (success only)       & \xmark \\
    GuessWhat?!~\cite{vries_guesswhat_2017}  & Target secrecy                    & \xmark & \xmark & Partial (guess success)      & \xmark \\
    TEACh~\cite{padmakumar_teach_2021}            & Oracle knowledge (Commander)      & \xmark & \xmark & \cmark                       & \xmark \\
    ALFRED~\cite{shridhar_alfred_2020}          & None (generic listener, one-way)  & \xmark & \xmark & \cmark\ (task success)       & \xmark \\
    DialFRED~\cite{gao_dialfred_2022}      & Agent-initiated clarification     & \xmark & \xmark & \cmark\ (task success)       & \xmark \\
    RefCOCO family~\cite{kazemzadeh_referitgame_2014,mao_generation_2016} & None (single grounding model) & \xmark & \xmark & Partial (grounding accuracy) & \xmark \\
    NR3D / ReferIt3D~\cite{achlioptas_referit3d_2020} & Difficulty (viewpoint, distractors) & \xmark & \xmark & Partial (accuracy only) & \xmark \\
    \midrule
    \textbf{Ours}                 & \textbf{\shortstack[l]{Manipulated listener knowledge\\(task, robot-part, scene)}}
      & \cmark & \cmark & \cmark\ (success, time, moves)
      & \cmark\ (Workload, comprehension, usefulness) \\
    \bottomrule
  \end{tabular}}
  \label{tab:table1} 
\end{table*}

\subsection{Knowledge Differences in Human-Centered Explanations}
HCI research on explainable systems shows that explanations must account for the person receiving them. Liao et al.~\cite{liao2020questioning} identified gaps between technical approaches to explainability and the questions that users bring to AI systems. Schaffer et al.'s IUI study~\cite{schaffer2019expertise} found that the effect of explanations differed with reported task familiarity. These studies motivate treating user knowledge as part of communication design. LD-HRI differs by assigning access to task information, holding the request constant, and measuring whether the Listener completes the requested action.


\section{LD-HRI Platform and Data Collection}
    We designed this study to measure human performance given a robot's help request while controlling for differences in a listener's knowledge. Unlike prior work~\cite{takmaz_speaking_2023}, which uses LLMs as listeners to test HRI task performance, we instead adapted robot-help request scenarios to the structure of communication games such as the Cards Corpus \cite{potts_goal-driven_2012}. We recruit human participants as listeners, and observe a listener's performance based on the same human-authored help request provided to them, and the different visual information available to expert and novice listeners. Our method aims to obtain a dataset that investigates other human differences in perspective-taking beyond viewpoint. 

\subsection{Design Goals}
We use three design goals to isolate whether a robot's first help request contains the information needed by Listeners with different task information.

\textbf{DG1: Control the information available to the Listener for standardization purposes.}
Expert and Novice conditions are defined by information shown in the interface, not self-reported expertise, as participants' estimation of their own expertise
might be hard to standardize. This provides a consistent basis for comparing listeners with different task information.

\textbf{DG2: Separate forms of familiarity beyond viewpoint.}
Prior work focused on investigating human differences with respect to their viewpoints. However, humans differ in other ways beyond viewpoints, such as their understanding of the environment and robots.
We manipulate control-interface, robot-part, and scene information. Tasks 1--3 isolate one form at a time, while Task 4 combines scene and robot-part information. We do not compare which familiarity type is harder because the tasks also differ in structure and the potential for listeners to discover information by exploration.

\textbf{DG3: Isolate the information carried by the first help request.}
In the real world, a human may need to act immediately or decide to do so without clarification, making the initial wording of the request critical to how the responder comprehends and acts on the request. If clarification were allowed, an incomplete first request could be corrected during the exchange, making it difficult to judge the original wording.
LD-HRI evaluates a one-shot initial request before clarification or repair. This allows the same authored request to be given to Listeners with different information and tests whether the first message contains enough information for successful assistance.

\subsection{LD-HRI Game Design}
Each round has a Speaker and a human Listener. The Speaker sees the robot's goal and task reference information and writes one request that should enable the Listener to help. The Listener reads the request and acts in the game to complete that goal: drive to a location, connect parts, retrieve an object, or navigate and then repair. The Listener cannot ask the Speaker for clarification. Success is determined by the resulting task state, not by a self-report. Expert and Novice Listeners pursue the same goal but receive different reference information.

We designed the Listener Differences in Human-Robot Interaction (LD-HRI) game because no existing dataset provides the same human-authored robot help request to people performing the same task under experimentally controlled listener information. The game holds the Speaker request fixed while varying the visual reference information available to the Listener and records human actions, outcomes, workload, and perceptions.

We use human Speakers and Listeners because there is a need to understand the effectiveness of different
speaker utterances on actual humans instead of relying on models as human proxies.  LD-HRI lets us benchmark human differences using real human responses instead of defining Listener variation solely through computational models. The resulting paired corpus links each served Speaker request to Expert and Novice outcomes where both are available.

\begin{table*}[t]
  \centering
  \caption{\textbf{Listener information manipulated across the four LD-HRI tasks.} $N$ reports the number of requests with both Expert and Novice outcomes used in the matched analysis.}
  \Description{Table listing, for each of the four LD-HRI tasks, what the Listener does, what the Expert has, what the Novice lacks, the familiarity type manipulated, and the number of matched requests.}
  \label{tab:tasks}
  \small
  \resizebox{\textwidth}{!}{%
  \begin{tabular}{@{}l l l l l r@{}}
    \toprule
    \textbf{Task} & \textbf{Listener does} & \textbf{Expert has} & \textbf{Novice lacks} & \textbf{Familiarity} & \textbf{Matched $N$} \\
    \midrule
    1. Teleoperation & Drive robot to hidden goal & Control legend & Control legend & Robot control interface & 91 \\
    2. Hardware Repair & Connect two robot parts & Part labels & Part labels & Robot part & 84 \\
    3. Navigate then Retrieve Object & \shortstack[l]{Explore building, find and\\pick up target} & \shortstack[l]{Room and object labels\\from start} & \shortstack[l]{Room/object labels initially;\\room name appears after entry} & Scene & 89 \\
    4. Integrated Navigation and Repair & \shortstack[l]{Navigate to failure room, identify\\two parts, connect them} & Room and part labels & Room and part labels & Scene + robot part & 57 \\
    \bottomrule
  \end{tabular}}
\end{table*}

\subsection{Task Design}
\label{sec:game}

All four tasks follow this request--action structure. The first three vary control, part, or scene information; Integrated Navigation and Repair combines scene and part information. Each scene has an objective goal and a fixed scoring rule. Integrated Navigation and Repair was added later and occupied the final task position; task-specific comparisons therefore also reflect differences in task structure and position. 

\begin{description}
    \item[Task 1. Teleoperation:] Task 1 models control-interface familiarity by asking the Listener to drive a robot to a hidden goal. The Speaker sees the goal and key-to-motion mapping. The Expert Listener sees the control mapping, while the Novice does not. A request containing only directions may therefore work for the Expert, while a Novice may need the corresponding keys or must learn the mapping through exploration.

    \item[Task 2. Hardware Repair:] Task 2 models robot-part familiarity by asking the Listener to identify and connect two components. The Speaker and Expert see fictional technical labels such as ``Vornak'' and ``Torvin,'' while the Novice sees the same components without labels and must rely on descriptions such as color, shape, or location. Since the fictional name-to-part mapping cannot be discovered through exploration, this task tests whether the request provides another way to identify the intended components.

    \item[Task 3. Navigate then Retrieve Object:] Task 3 models scene familiarity by asking the Listener to navigate through a building and retrieve a target object. The Speaker sees the full layout, room names, object information, and target. Listeners see only their current room, and room names appear after entry. This design is informed by spatial referring-expression benchmarks such as ReferIt3D~\cite{achlioptas_referit3d_2020}. Unlike Hardware Repair, some missing scene information can therefore be recovered through exploration.

    \item[Task 4. Integrated Navigation and Repair:] Task 4 combines navigation and robot-part identification. The Listener navigates to the robot's failure room and connects two relevant parts. The Expert sees room and part names, while those labels are withheld from the Novice. Two matched layouts repeat the manipulation. Since scene and part information are withheld together, the task can identify whether failure occurred during navigation or part connection but cannot causally separate the two familiarity types.
\end{description}
\subsection{Task Platform Setup and Logging}
LD-HRI is a browser-based Next.js application backed by Postgres and deployed through Vercel. The server assigns roles, exposes only the information appropriate to each condition, stores Speaker requests, and validates Listener actions without exposing hidden task state. The platform records requests, actions, timestamps, state changes, outcomes, workload, and Listener ratings so each trial can be reconstructed and linked to its Speaker request and Listener condition.

Participants completed two layouts of each original task. Listeners had three minutes per trial. Hardware Repair allowed four connection attempts, Teleoperation 75 key presses, and Object Retrieval 45 moves. Integrated Navigation and Repair allowed 60 navigation moves, four connection attempts, and four minutes. A trial ended earlier on success. Speakers had no time limit because the study evaluates request usefulness instead of writing speed.

\subsection{Participants}
\begin{figure}[t]
  \centering
  \includegraphics[width=\linewidth]{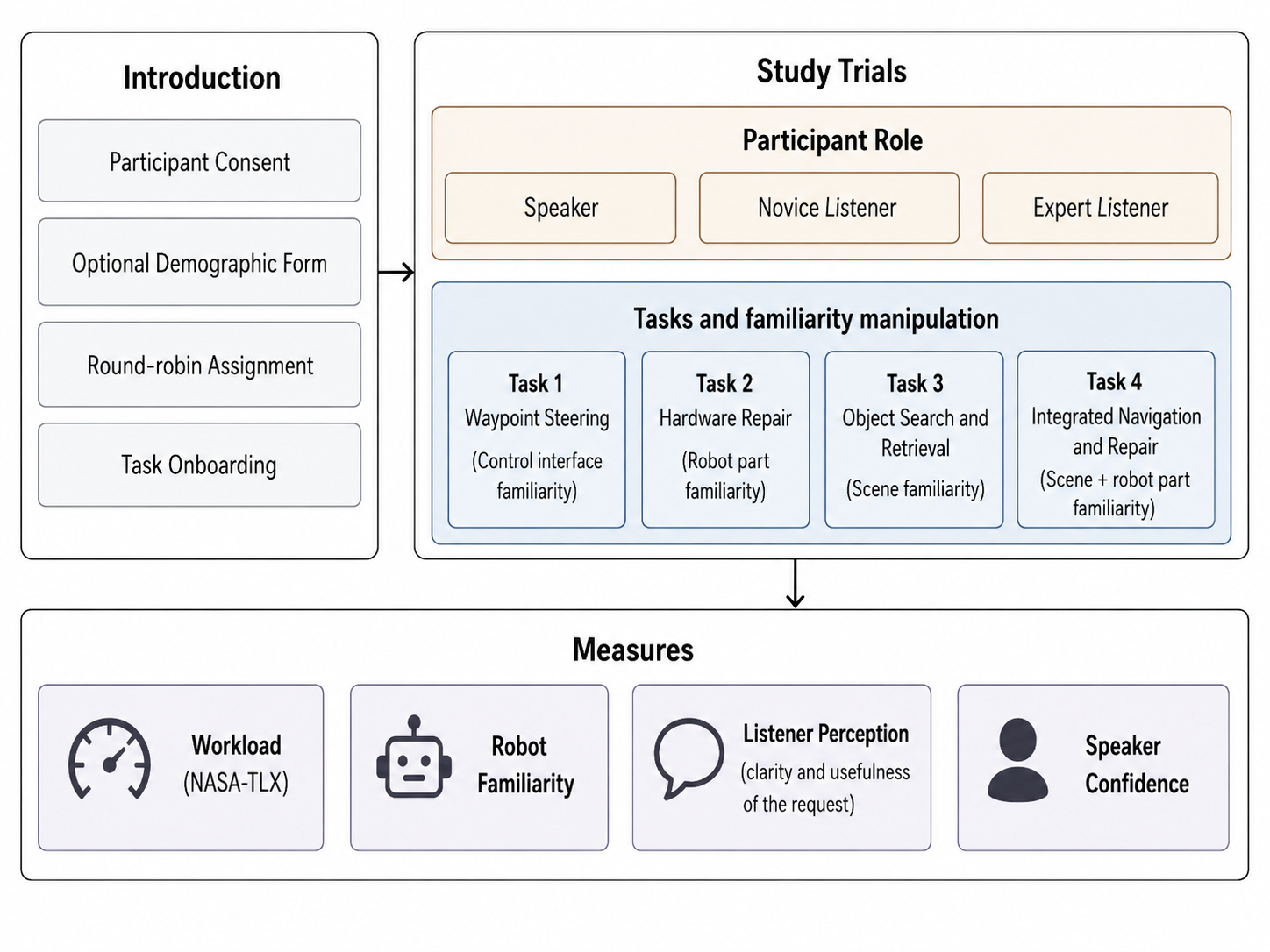}
  \caption{LD-HRI study procedure across four tasks manipulating control interface, robot part, and scene familiarity, with workload, robot familiarity, Listener perceptions, and Speaker confidence recorded.}
  \Description{Flow diagram of the LD-HRI study procedure across the four tasks, showing the recorded measures.}
  \label{fig:procedure}
\end{figure}

We recruited 240 participants through word of mouth and posting on social media under an approved institutional review board protocol over three months.
Participants were recruited in rolling groups with a target ratio of five Speakers, five Novice Listeners, and five Expert Listeners. The original three-task cohort was approximately balanced, with 25 Speakers, 25 Novice Listeners, and 24 Expert Listeners. During the later data collection that included the Integrated Navigation and Repair task, completed role counts became unbalanced, resulting in 37 Speakers, 100 Novice Listeners, and 29 Expert Listeners in that cohort.
The final dataset therefore contains 62 Speakers, 125 Novice Listeners, and 53 Expert Listeners. Our primary Expert--Novice analysis does not compare these unequal participant totals directly. Instead, it uses the 321 Speaker requests with observed outcomes from both Listener conditions, holding the request wording fixed.
The released data retains all authored Speaker requests, while analyses of Listener behavior use only requests that were actually presented to a Listener. See procedure details in \autoref{fig:procedure}.

We collected demographics and prior robot familiarity to describe the study population, not to determine whether a participant was assigned to the Expert or Novice condition. Participants were compensated \$7 for approximately 25 minutes of participation. Prior robot familiarity was measured on a five-point scale from 1 (very unfamiliar) to 5 (very familiar). The familiarity item has 219 valid responses, as 21 participants were skipped during earlier pilots. We do not infer or recode values for those 21 participants. Among the 219 valid responses, the mean familiarity rating was 3.24 (median = 3), and 172 participants (78.5\%) rated their familiarity as 3 or higher. Mean familiarity was 3.11 for Expert Listeners and 3.36 for Novice Listeners. We found no evidence that the familiarity distribution differed between Expert and Novice Listeners ($p=.470$).
Participants reported age range, gender, race or ethnicity, and field of study or work. Demographic questions were optional initially and included a preferred not-to-say option. However, we reached out to participants to collect the demographic information before fund disbursement.
We control our experimental Experts and Novices conditions via access to visual information shown in the LD-HRI interface.  Table~\ref{tab:participants} summarizes the participant demographics, characteristics, prior familiarity with the robot, and compensation.

\begin{table*}[t]
  \centering
  \caption{\textbf{Participant characteristics for the 240 completed study sessions.}
  Demographic questions were optional, so demographic category counts do not necessarily sum to 240. Robot familiarity was measured from 1 (very unfamiliar) to 5 (very familiar) and was available for 219 participants.}
  \Description{Table summarizing participant roles, age, gender, race or ethnicity, field of study or work, prior robot familiarity, and compensation for the 240 completed participants.}
  \label{tab:participants}
  \small
  \begin{tabularx}{\textwidth}{@{}l X@{}}
    \toprule
    \textbf{Characteristic} & \textbf{Participant information} \\
    \midrule

    Completed participants &
    240 total: 62 Speakers, 53 Expert Listeners, and 125 Novice Listeners. \\

    Age &
    Participants ranged from 18 to 64 years, 129 participants were in the 18--24 age range. \\

    Gender &
    Man: 160, Woman: 52. \\

    Race or ethnicity &
    Black or African American: 126, Asian: 40, White: 17, Middle Eastern or North African: 9, preferred not to report: 30. \\

    Field of study or work &
    Among participants reporting a field, 65\% reported computing, engineering, or robotics. \\

    Prior robot familiarity &
    219 valid responses; mean = 3.24, median = 3. Of these, 172 (78.5\%) rated their familiarity as 3 or higher. Mean familiarity was 3.09 for Speakers, 3.11 for Expert Listeners, and 3.36 for Novice Listeners. \\

    Compensation &
    \$7 for approximately 25 minutes of participation. \\

    \bottomrule
  \end{tabularx}
\end{table*}

\subsection{Data Collection and Dataset Construction}

Across 240 completed study sessions, 62 Speakers authored 446 robot help requests. Of these, 395 requests were presented to at least one human Listener, producing 1{,}302 Listener trials across the four tasks. A total of 321 requests have observed outcomes from both an Expert and a Novice Listener and form the paired dataset used for the primary matched analyses. Each Listener trial links the original Speaker request to the Listener information condition, task and scene, task outcome, recorded actions, timestamps, workload responses, and Listener ratings where available. This linkage allows the same human-authored request to be compared across Expert and Novice Listener conditions while holding the request wording fixed. We provide the LD-HRI dataset, game platform, and analysis code to support replication and future evaluation of listener-aware robot communication.

\subsection{Study Design}
The study uses three between-subjects roles, Speaker (C$_1$), Expert Listener (C$_2$), and Novice Listener (C$_3$). Each participant completes one role through round-robin assignment. Expert and Novice refer to task-specific information provided by the interface rather than prior expertise. When the same Speaker request reaches both Listener conditions, we compare their outcomes while holding the wording fixed. Integrated Navigation and Repair was added later. We report its task-specific result separately, while the overall dataset summaries and matched RQ1 comparison use the available data across all four tasks.

\subsubsection{Roles and Information Access}
The Speaker sees the task goal and reference information needed to describe the correct action, paralleling a robot that knows its failure state but cannot complete the recovery action itself~\cite{tellex_asking_2014}. Listeners can act in the task but receive only the reference information available to their assigned condition. Experts receive the task-specific information shown in Table~\ref{tab:tasks}. Novices perform the same task with that information withheld. Both conditions receive the same authored request, separating message content from listener information.

In Tasks 1--3, Speakers are told what information a Listener may lack but not which Listener condition will receive the message. In Task 4, Speakers are told that the same message will go to both conditions (Table~\ref{tab:speaker-knowledge}).

\begin{table*}[t]
  \centering
  \caption{\textbf{Information available to Speakers, Expert Listeners, and Novice Listeners across LD-HRI tasks.}}
  \Description{Table showing the task information available to the Speaker, Expert Listener, and Novice Listener in each LD-HRI task.}
  \label{tab:speaker-knowledge}
  \small
  \resizebox{\textwidth}{!}{%
  \begin{tabular}{@{}l l l l@{}}
    \toprule
    \textbf{Task} & \textbf{Speaker information} & \textbf{Expert Listener} & \textbf{Novice Listener} \\
    \midrule
    Teleoperation &
    \shortstack[l]{Sees goal, route, and\\control mapping} &
    Sees control mapping &
    Control mapping withheld \\

    Hardware Repair &
    \shortstack[l]{Sees target parts and\\their technical labels} &
    Sees part labels &
    Part labels withheld \\

    Object Retrieval &
    \shortstack[l]{Sees full layout, room names,\\objects, and target} &
    \shortstack[l]{Receives task-specific\\scene information} &
    \shortstack[l]{Room/object labels initially withheld;\\room name appears after entry} \\

    Integrated Navigation and Repair &
    \shortstack[l]{Sees room and part information;\\told request goes to both conditions} &
    Sees room and part labels &
    Room and part labels withheld \\
    \bottomrule
  \end{tabular}}
\end{table*}

\begin{figure}[t]
  \centering
  \includegraphics[width=\linewidth]{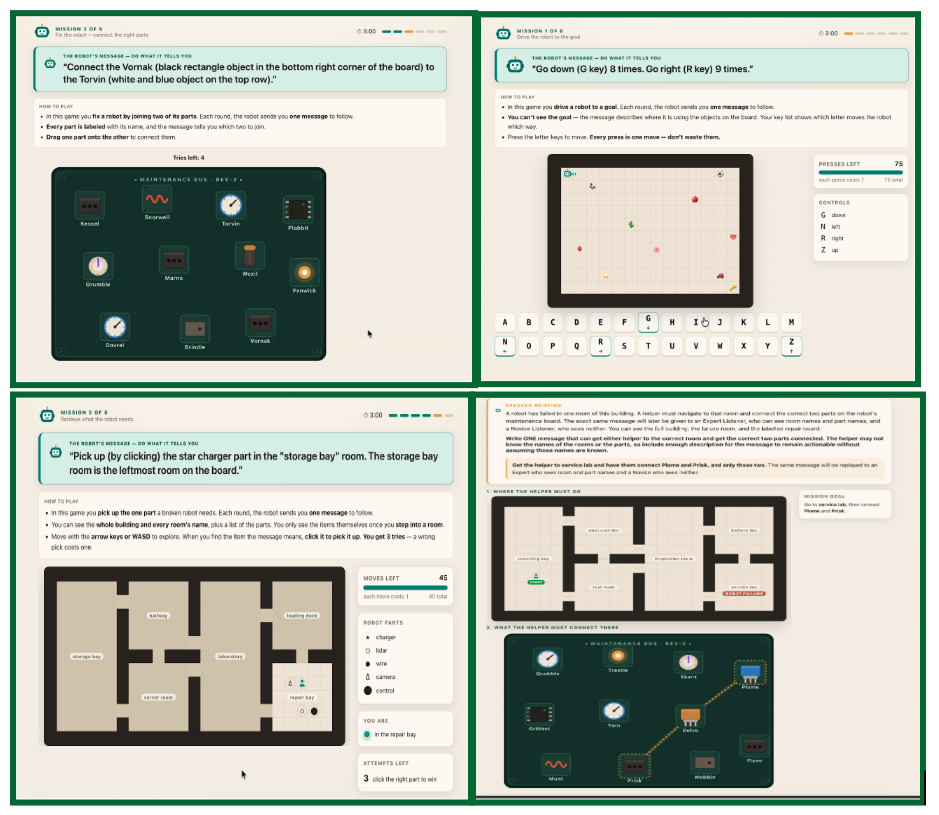}
  \caption{Expert Listener views for the three original tasks, Teleoperation, Hardware Repair, and Object Retrieval. Task 4 combines navigation and repair and is summarized in Table~\ref{tab:tasks}.}
  \Description{Expert Listener views for the three original tasks: Teleoperation, Hardware Repair, and Object Retrieval.}
  \label{fig:task-views}
\end{figure}

\subsection{Study Procedure}
Participants followed the same study sequence within their assigned role.
Participants entered through the study link, reviewed the consent form, and completed an intake survey containing robot familiarity, field of study or work, and optional demographics.
After role assignment, participants received the instructions for their role before beginning the trials.
Participants in the original study completed six trials, consisting of two versions of each of the three original tasks described in the Game section.
After each trial, participants completed the NASA Task Load Index (NASA-TLX) \cite{hart1988development}and the questions specific to their role. Speakers rated confidence in their request, while Listeners rated the usefulness of the request they received.
After all trials, participants could provide optional written feedback.

Study duration differed by role because Speakers wrote requests while Listeners acted on requests that had already been written.
Listener sessions had a median duration of 11.3 minutes, while Speaker sessions had a median duration of 28.6 minutes.
Speakers spent a median of 21.2 minutes actively composing requests across the six original-task trials.
The task-specific design and time limits have been rationalized in Section~\ref{sec:game}.

\subsection{Evaluation Metrics}
We record objective task outcomes, listener ratings, and speaker measures.

\begin{description}
\item[Task success.]
The game's scoring rules determine success from the task state. Questionnaire responses do not determine success.

\item[Time and task cost.]
We record trial duration and task-specific action costs, including movements in Teleoperation and Object Retrieval, connection attempts in Hardware Repair, and both in Integrated Maintenance. Room entries and failure reasons are recorded where available. Time and cost summaries include unsuccessful trials, so lower values alone do not establish greater efficiency.

\item[Listener workload.]
The NASA Task Load Index (NASA-TLX)~\cite{hart1988development} measures mental, physical, and temporal demand, effort, frustration, and perceived performance. Perceived performance is subjective and is separate from objective task success.

\item[Clarity and usefulness.]
Listeners rate the clarity and usefulness of each request. Response-scale changes and missing-value handling are described in the analysis.

\item[Speaker measures.]
We record the exact request, composition time, and the speaker's confidence on a five-point scale that the listener will complete the task. Confidence is separate from observed listener success.
\end{description}

\subsection{Analysis and Benchmark}
We organize the analysis around four research questions and use human listener outcomes to evaluate help requests.

For RQ1, we compare task success and workload across expert and novice information conditions. We also examine the association between perceived request clarity and workload.

For RQ2, we examine the association between referring strategies in human-written requests and listener success.

For RQ3, we evaluate LLM-generated requests with human listeners and compare their outcomes descriptively with the earlier human-speaker corpus.

For RQ4, we evaluate requests selected by the inverse-semantics reference policies $S_1$ and $S_2$ using observed human listener outcomes.

Table~\ref{tab:analyses} summarizes the units, samples, and analyses used for each evaluation.

\begin{table*}[t]
\centering
\caption{\textbf{Overview of the LD-HRI analyses and the data used for each evaluation.}}
\Description{Table listing each research question, its unit of analysis, analysis sample, and reported analyses.}
\label{tab:analyses}

\normalsize
\renewcommand{\arraystretch}{1.2}
\setlength{\tabcolsep}{4pt}

\begin{tabularx}{\textwidth}{@{}
>{\raggedright\arraybackslash}p{0.17\textwidth}
>{\raggedright\arraybackslash}p{0.17\textwidth}
>{\raggedright\arraybackslash}X
>{\raggedright\arraybackslash}p{0.23\textwidth}
@{}}
\toprule
\textbf{Evaluation} &
\textbf{Unit} &
\textbf{Analysis sample} &
\textbf{Analysis or report} \\
\midrule

RQ1. Task success &
Speaker request &
321 matched requests across four tasks &
McNemar's exact test and task-specific comparisons \\
\addlinespace

RQ1. Workload and perception &
Listener for workload.
Trial for clarity--workload association. &
164 listeners with NASA-TLX, including 46 expert and 118 novice listeners.
1{,}238 linked trial records.
Correlation uses records with both ratings available. &
Mann--Whitney test and Spearman correlation \\
\addlinespace

RQ2. Referring strategies &
Listener trial &
Human-written requests grouped by referring strategy and task &
Success by referring strategy and Fisher's exact test \\
\addlinespace

RQ3. LLM speakers &
Listener trial &
24 frozen requests across eight scenes.
70 human listeners and 560 trials. &
Expert and novice success, time, task cost, and workload \\
\addlinespace

RQ4. Reference policies &
Selected request within scene &
321 candidate requests with expert and novice outcomes across eight scenes &
Expert and novice selected-request success \\

\bottomrule
\end{tabularx}
\end{table*}

\subsubsection{Task Performance and Experience (RQ1)}
\label{sec:rq1-analysis}
\paragraph{Task success.}
To answer RQ1, we compare Expert and Novice outcomes for the same Speaker request.
For requests evaluated by both listeners, we pair the Expert and Novice outcomes for the same Speaker request. McNemar's exact test~\cite{fagerland_mcnemar_2013} then compares the requests where the two Listeners had different outcomes, e.g Expert success with Novice failure versus Novice success with Expert failure.
McNemar's exact test is appropriate because task success is binary and the two outcomes correspond to the same speaker request.
We report the Expert-Novice difference as Expert success minus Novice success in percentage points, overall and separately by task.
Task comparisons are descriptive because the tasks differ in more than the type of information withheld, including whether missing information can be discovered during the task.

We also check whether the RQ1 result holds for the same people who contributed to several trials.
Each Speaker contributes multiple requests, and each Listener completes multiple trials, so observations may be correlated within Speakers and within Listeners.
We therefore test the robustness of the RQ1 result using a permutation test that shuffles the Expert and Novice labels at the participant level, together with generalized estimating equation (GEE) logistic regression models, once accounting for repeated observations from the same Speaker and once for repeated observations from the same listener.
We compare the overall success rates of Expert and Novice Listeners using a Mann-Whitney test~\cite{mann_test_1947}. This avoids treating the six trials completed by one listener as six independent observations.
We report all of these checks so that the main result is not based on a single statistical test.

\paragraph{Sample-size sensitivity.}
We conducted a sensitivity analysis for the primary RQ1 matched comparison using a two-sided McNemar test with $\alpha=.05$ and 90\% power. The final matched dataset contains 321 requests with both Expert and Novice outcomes. Of these, 106 requests (33.0\%) have discordant outcomes. Experts succeeded when Novices failed on 80 requests, while Novices succeeded when Experts failed on 26. Expert Listeners succeeded on 76.9\% of matched requests compared with 60.1\% for Novice Listeners, an Expert--Novice difference of 16.8 percentage points. The separate LLM-speaker follow-up is described in Section~\ref{sec:llm-speaker-method}.

Given the observed 33.0\% discordance rate and 16.8 percentage-point difference, approximately 123 matched requests would be required to achieve 90\% power. The final dataset contains 321 matched requests, approximately 198 more than this requirement. As an additional sensitivity check, with 321 matched requests and the observed discordance rate, the study has approximately 90\% power to detect an Expert--Novice difference of about 10.4 percentage points. These results indicate that the final matched sample is adequate for the primary RQ1 comparison.

\paragraph{Listener workload and perception.}
To answer RQ1, we test whether workload differs between Expert and Novice Listeners and whether clearer requests are associated with lower workload.
For each Listener with available NASA Task Load Index (NASA-TLX) data, we calculate their mean workload across completed trials so that each Listener contributes one value to the overall condition comparison. The dataset contains 1{,}238 linked NASA-TLX trial records from 46 Expert and 118 Novice Listeners. We compare the participant-level mean workload of Expert and Novice Listeners using a Mann--Whitney test~\cite{mann_test_1947}, which does not assume normally distributed workload scores. We report the overall NASA-TLX score and the six individual subscales across mental demand, physical demand, temporal demand, perceived performance, effort, and frustration. For the six subscale comparisons, we apply the Benjamini--Hochberg correction~\cite{benjamini_controlling_1995} to control for multiple comparisons.

We separately examine how Listener perceptions of the request relate to workload.
Listeners rated the clarity and usefulness of the request after each trial.
We use Spearman correlation~\cite{spearman_proof_1904} to evaluate the relationship between clarity and NASA-TLX separately for Expert and Novice Listeners, allowing us to assess monotonic relationships without assuming normally distributed data.
We use Fisher's r to z test to compare whether the relationship between clarity and workload differs between experts and novices.
Fisher's test converts correlation coefficients into normally distributed z scores to let us calculate confidence intervals.
Completion time, actions taken, rooms entered, and reasons for failure are reported descriptively.

\subsubsection{Referring Strategies (RQ2)}
\label{sec:rq2-analysis}
To answer RQ2, we first test if the speaker's description of task information is associated with listener success.
In Hardware Repair, we classify requests according to whether they use robot part names, visible descriptions such as color and shape, or both.
We use Fisher's exact test to test whether including a visible description is associated with success separately for Expert and Novice Listeners.
Fisher's exact test is used because some of the groups contain only a small number of requests.
We apply the same analysis to Object Retrieval when enough requests are available for comparison.
Teleoperation contains too few requests with comparable visible descriptions for the same analysis, so we do not make a statistical comparison for that task.

\subsubsection{Evaluating LLM Speakers with Human Listeners (RQ3)}
\label{sec:llm-speaker-method}
We evaluate GPT-5.6 Sol, Claude Sonnet 5, and Claude Haiku 4.5 as Speakers. Each receives the task information available to a human Speaker and generates one request for each of eight scenes, yielding 24 frozen requests. For each model and scene, the same request is presented to participants in both the expert and novice listener conditions.
The follow-up contains 70 human Listeners, 35 Expert and 35 Novice. Each participant is assigned to one Speaker model and completes eight trials, two per task, giving 560 trials. Expert/Novice participant counts are 12/11 for GPT, 11/12 for Sonnet, and 12/12 for Haiku. These counts describe the follow-up sample; participants are not assumed to be distinct from all earlier recruits.

We score success using the game's objective task rules and report elapsed trial time, recorded task cost, and human questionnaire ratings separately. Task cost counts movements in Teleoperation and Object Retrieval, connection attempts in Hardware Repair, and both in Integrated Maintenance. It is not a universal count of interface actions. NASA-TLX ratings are subjective and do not determine task success.

We report success rates for each LLM separately. For each task, we also combine results from the three LLMs and compare them descriptively with results for human-written requests from the original study. These studies differ in collection and sample composition, so the Speaker-type comparisons are descriptive. The current analysis does not test whether an LLM can adapt its request when explicitly told the intended audience.
\subsubsection{Evaluating Reference Policies (RQ4)}
\label{sec:rq4-analysis}
We adapt the $S_0$, $S_1$, and $S_2$ inverse-semantics algorithms from Tellex et al.~\cite{tellex_asking_2014} to LD-HRI. The original $S_0$ uses the fixed request ``Help me'' and models neither the environment nor the listener. This exact request was not presented to human listeners in LD-HRI, so we include $S_0$ as the fixed baseline but do not assign it a retrospective human success rate.
For $S_1$, we score each candidate using task information expressed in the request, without using human listener outcomes. This represents an environment-aware policy that does not model differences between listeners. For $S_2$, we use one general Listener-success model trained on both Expert and Novice trials without providing the Listener condition to the model.

For the retrospective comparison, we restrict candidate requests to the 321 requests with observed outcomes from both an Expert and a Novice Listener across eight task scenes. Within each scene, $S_1$ and $S_2$ select the candidate with the highest corresponding score.

\section{Results}

\subsection{Descriptive Statistics}

The analyzed dataset contains 62 Speakers, 178 Listeners, and 1{,}302 Listener trials across four tasks.
The 62 analyzed Speakers authored 446 help requests, and 395 were presented to
at least one Listener. A total of 321 requests have outcomes from both an
Expert and a Novice Listener and are used for the matched comparison. Across
all 1{,}302 Listener trials, Experts succeeded on 278 of 365 trials (76.2\%),
while Novices succeeded on 449 of 937 trials (47.9\%). Figure~\ref{fig:matched-pairs} summarizes the matched outcomes.

\subsection{Task Performance and Experience (RQ1)}
\begin{figure}[!htbp]
  \centering
  \includegraphics[width=\linewidth]{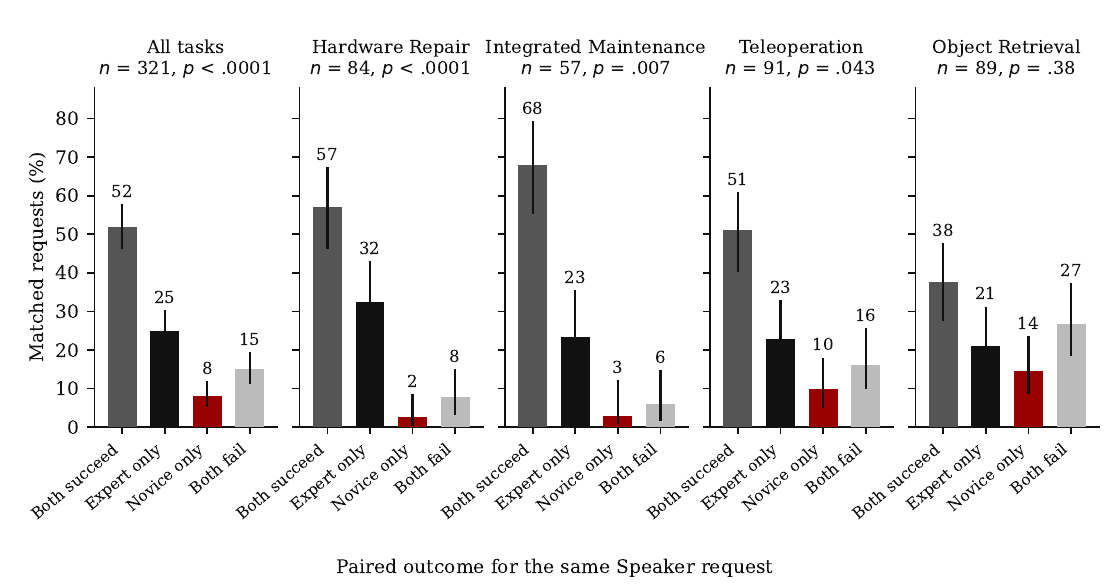}
  \caption{Matched Expert and Novice outcomes overall and by task. Bars show the paired outcomes for the same Speaker request; error bars show 95\% Wilson confidence intervals.}
  \Description{Bar chart of the four paired Expert and Novice outcomes for the same Speaker request, overall and by task, with 95 percent Wilson confidence intervals.}
  \label{fig:matched-pairs}
\end{figure}

\paragraph{Task success.}
Figure~\ref{fig:matched-pairs} summarizes expert and novice outcomes for the same requests. Across 321 requests evaluated in both conditions, 255 had exactly one expert and one novice outcome, while 66 had multiple outcomes in at least one condition. We selected one observed outcome per condition for each request and repeated the pairing 2{,}000 times to check sensitivity to the choice of outcomes.

Across these pairings, an average of 166.6 requests had both listeners succeed and 48.5 had both fail. Expert success with novice failure was more common than the reverse, with 80.3 versus 25.7 requests on average. In a verification run, McNemar's exact test gave $p<.0001$ for every pairing. This checks sensitivity to pairing choices and does not account for repeated observations from the same participants.

We tested whether the Expert advantage remains after accounting for repeated observations from the same Speakers and Listeners.
A permutation test that keeps each listener's trials together gives $p=.0106$.
Generalized estimating equation (GEE) logistic regression model accounting for multiple requests written by the same Speaker gives odds ratio (OR) = 2.05, $p = .013$. Hence, estimated odds of success are 2.1 times higher for experts than novice listeners with statistical significance.
The corresponding GEE model accounting for multiple trials completed by the same Listener gives OR = 1.96, $p = .0294$.
Comparing each Expert and Novice Listener's overall success rate also does not reach significance (Mann-Whitney U = 550, $p = .0759$).
We therefore interpret the Expert advantage as well-supported for matched requests and across Speakers, but we do not find a significant difference when comparing each Listener's overall success across all of their trials.

The difference between Expert and Novice outcomes is largest in Hardware Repair and smallest in Object Retrieval.
Among 84 matched Hardware Repair requests the odds ratio is 12.9 ($p <
.0001$). Among 57 matched Integrated Maintenance requests it is 8.0 ($p =
.007$). Among 91 matched Teleoperation requests it is 2.3 ($p = .043$). Among
89 matched Object Retrieval requests it is 1.4 and does not reach significance
($p = .38$). 
We do not use these task differences to conclude that one type of familiarity matters more than another, since the tasks also differ in how they are completed and what information can be recovered during play.

\begin{figure}[!htbp]
  \centering
  \includegraphics[width=\linewidth]{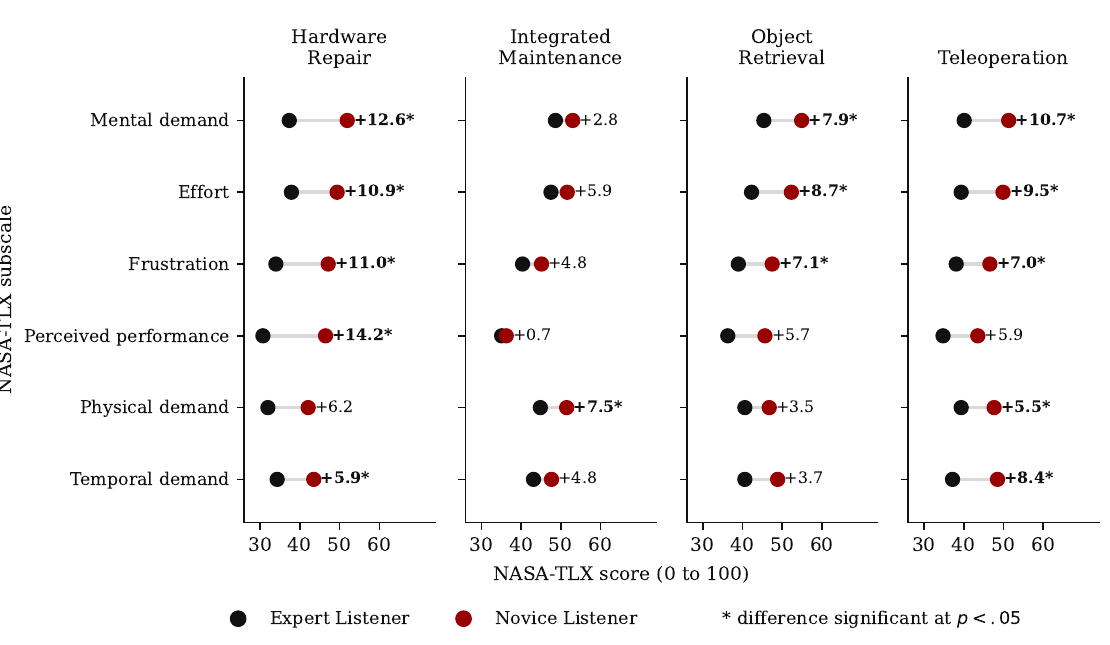}
  \caption{\textbf{Reported workload by Listener condition.} Higher scores indicate greater demand. Points show descriptive condition means by task and NASA-TLX subscale across 1{,}238 linked trial records. Overall and subscale significance tests use participant-level means as described in the Methods.}
  \Description{Chart of NASA-TLX subscale means by listener condition, with one standard deviation bars.}
  \label{fig:workload}
\end{figure}

\paragraph{Listener workload and perception.}
Figure~\ref{fig:workload} shows workload by task and listener condition. Novice Listeners reported significantly higher workload than Expert Listeners. After averaging NASA-TLX responses within each Listener, Novices reported a mean overall workload of 47.4 compared with 37.7 for Experts (Mann--Whitney $U=1522$, $p<.0001$). The analysis includes 118 Novice and 46 Expert Listeners.

Novice means were also higher on all six NASA-TLX subscales, and all six differences remained statistically significant after Benjamini--Hochberg correction.

Requests that Listeners rated as clearer were linked to lower workload in both Listener conditions.

The correlation between clarity and workload was $\rho=-.68$ for Novice Listeners and $\rho=-.60$ for Expert Listeners. Although the negative relationship was numerically stronger for Novices, the difference between the two correlations was not statistically significant (Fisher $z=-1.08$, $p=.282$).

\subsection{Referring Strategies and Listener Success (RQ2)}
\begin{figure}[!htbp]
  \centering
  \includegraphics[width=\linewidth]{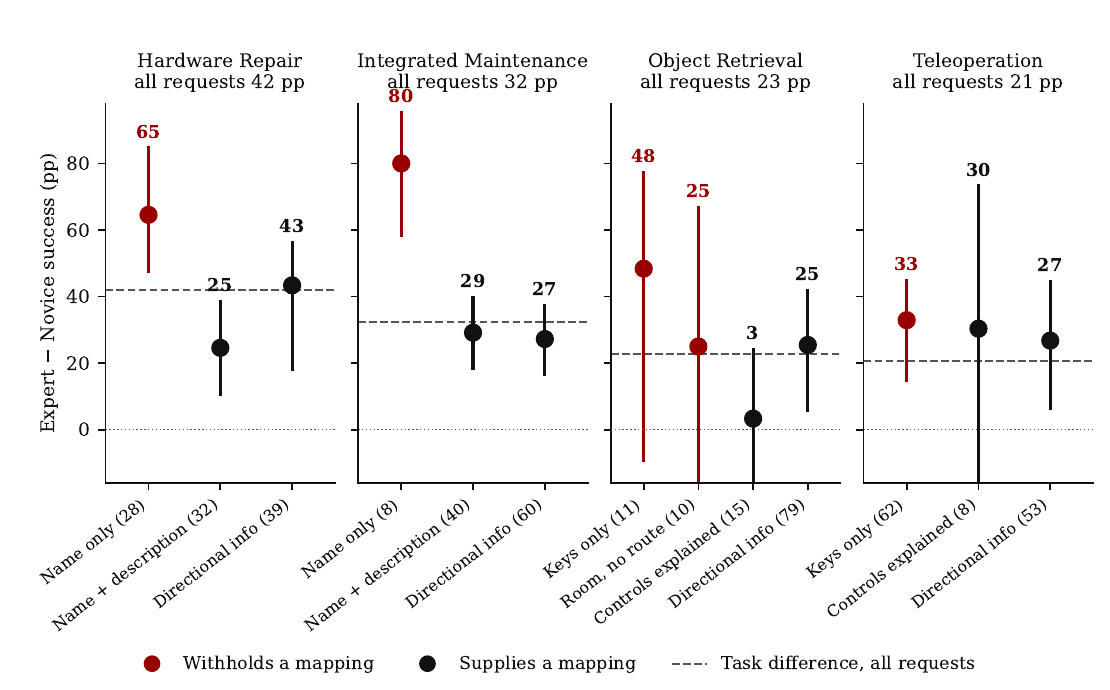}
  \caption{ Within every task, requests that withhold a mapping produce the largest Expert-Novice
difference. The dashed line is the difference across all requests in that task. Each task admits
only the properties that can occur in it. Properties resting on fewer than 15 requests carry wide
intervals and are reported for completeness.}
  \Description{Expert minus novice success by request property and task, with each task compared against its overall gap.}
  \label{fig:new_task}
\end{figure}

\begin{figure}[!htbp]
  \centering
  \includegraphics[width=0.9\linewidth]{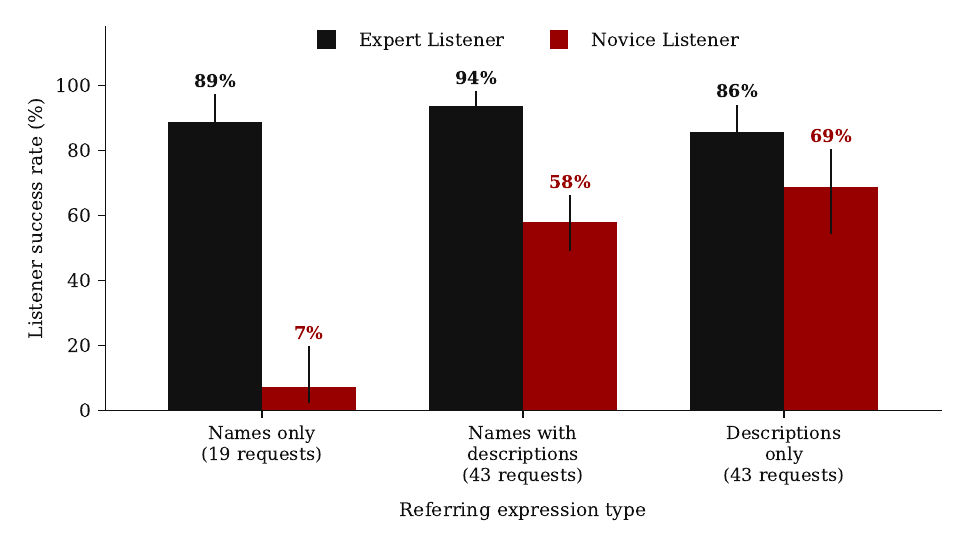}
  \caption{Hardware Repair success by referring strategy and Listener condition. Error bars show 95 percent Wilson confidence intervals.}
  \Description{Bar chart of Hardware Repair success rates by referring strategy for Expert and Novice Listeners.}
  \label{fig:referring-strategy}
\end{figure}

Figure~\ref{fig:new_task} provides a descriptive summary of task-specific request properties and expert--novice success differences.
Figure~\ref{fig:referring-strategy} compares referring strategies in Hardware Repair.
Hardware Repair provides the clearest example of how information included in a request relates to Listener success. Requests using only robot-part names succeeded for 88.9\% of Expert Listeners and 7.3\% of Novice Listeners across 19 requests. Requests combining names with visible descriptions succeeded for 93.6\% of Experts and 57.9\% of Novices across 43 requests. Requests using visible descriptions without part names succeeded for 85.7\% of Experts and 68.8\% of Novices across 43 requests. 

In Object Retrieval, Novice Listeners succeeded more often when requests included visual descriptions (82\% vs. 50\%), although the difference was not statistically significant ($p=.128$).
We did not test the association between visible descriptions and Teleoperation success because too few comparable requests were available.
These are observed associations and do not establish that adding visual descriptions causes an improvement in success.

\subsection{Human and LLM Speakers Leave an Expert--Novice Gap (RQ3)}
\label{sec:llm-speakers-human}
All Listeners in this evaluation are human. Human and Model in the figures identify who wrote the request. Across the 24 frozen LLM requests, Experts succeeded on 207 of 280 trials (73.9\%) and Novices on 161 of 280 (57.5\%), an observed gap of 16.4 percentage points. Thus, the evaluated models produce usable requests in many trials but do not eliminate the difference associated with access to reference information. Table~\ref{tab:llm-speaker-human-results} reports success separately for each model.

\begin{table}[!htbp]
\centering
\caption{LLM Speaker requests evaluated by human Listeners. Entries show successful/total trials and success percentages. These model comparisons are descriptive.}
\label{tab:llm-speaker-human-results}
\begin{tabular}{lrr}
\toprule
Speaker & Expert humans & Novice humans \\
\midrule
GPT-5.6 Sol & 71/96 (74.0\%) & 51/88 (58.0\%) \\
Claude Sonnet 5 & 68/88 (77.3\%) & 49/96 (51.0\%) \\
Claude Haiku 4.5 & 68/96 (70.8\%) & 61/96 (63.5\%) \\
\bottomrule
\end{tabular}
\end{table}

\begin{figure*}[!htbp]
\centering
\includegraphics[width=0.94\textwidth]{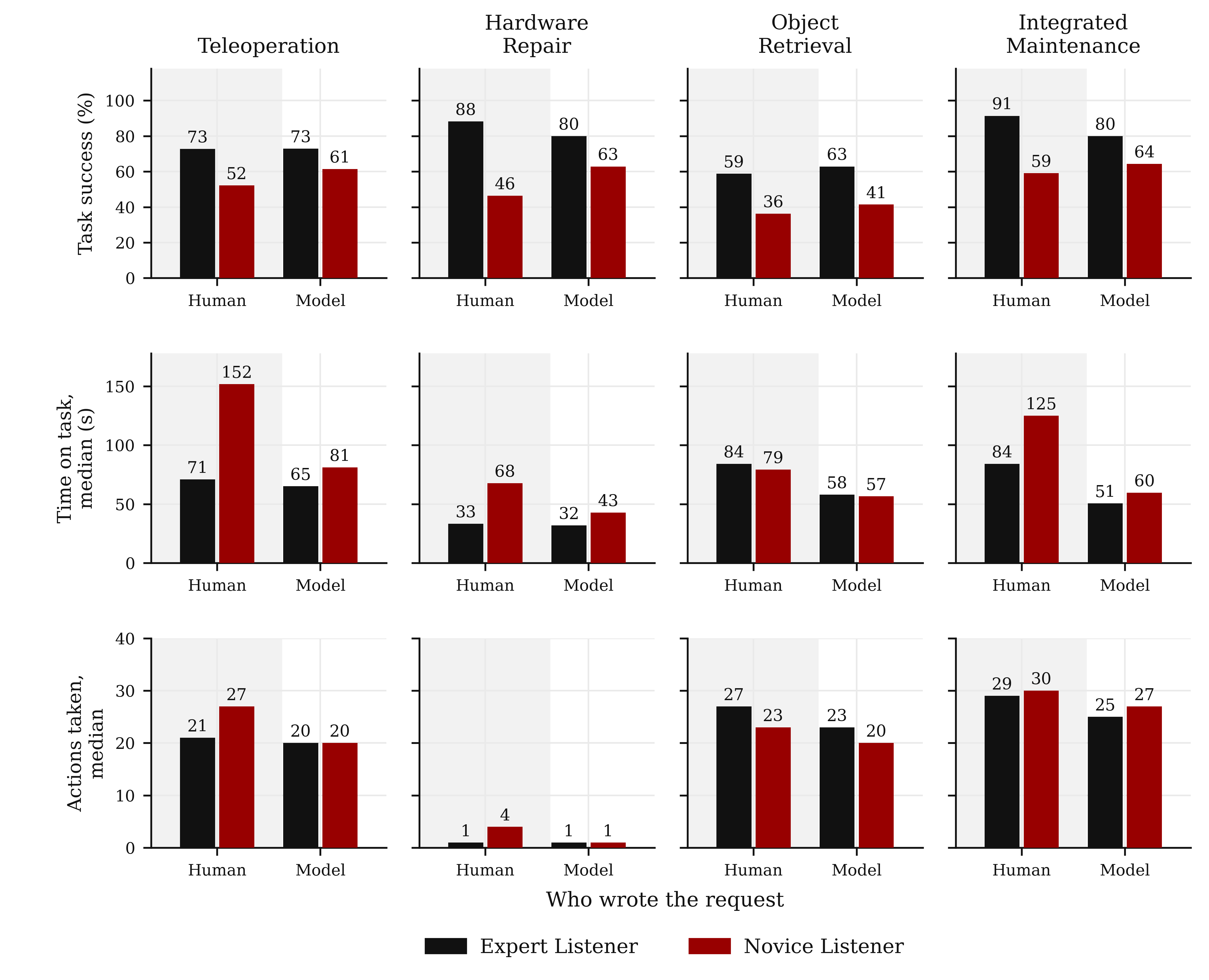}
\caption{\textbf{Human Listener outcomes for human- and model-written requests.} Model pools the three LLM Speakers. all Listeners are human. The shaded pair is the earlier human-Speaker reference. Time and task cost are medians over trials, including failures. Cost has a task-specific meaning. These comparisons describe separately collected studies rather than a controlled effect of Speaker type.}
\Description{Twelve panels compare Expert and Novice human outcomes for human and pooled model Speakers. Columns are four tasks; rows are objective success, median elapsed time, and median recorded task cost.}
\label{fig:outcome-by-task}
\end{figure*}

\paragraph{Outcomes across tasks.}
Figure~\ref{fig:outcome-by-task} compares human-written requests with the pooled model-written requests. Novice success is descriptively higher in the model study in every task: approximately 52\% versus 61\% in Teleoperation, 46\% versus 63\% in Hardware Repair, 36\% versus 41\% in Object Retrieval, and 59\% versus 64\% in Integrated Maintenance. Expert differences vary across tasks. The comparison does not establish that changing the speaker alone improves performance, because the studies were collected separately.

The Expert--Novice gap is seen across model-written requests in every task, ranging from 11.4 percentage points in Teleoperation to 21.4 in Object Retrieval. Retrieval also has the lowest pooled success in both model-study conditions (62.9\% Expert; 41.4\% Novice). This illustrates the challenge measured by the benchmark. Neither human-written requests nor these LLM-written requests reliably provide everything a less-informed Listener needs. Hence, methods needs to be designed to improve Novice success while maintaining Expert success.

\paragraph{Time and interaction cost.}
Novice median time was lower in the model study in all four tasks. For example, Teleoperation is 81 seconds versus 152 in the human corpus, and Integrated Maintenance is 60 versus 125. In Hardware Repair, median Novice connection attempts are one in the model study versus four in the human corpus. These summaries include failures, so shorter time or fewer actions should not be interpreted as more efficient successful completion.

\begin{figure*}[!htbp]
\centering
\includegraphics[width=0.94\textwidth,height=0.70\textheight,keepaspectratio]{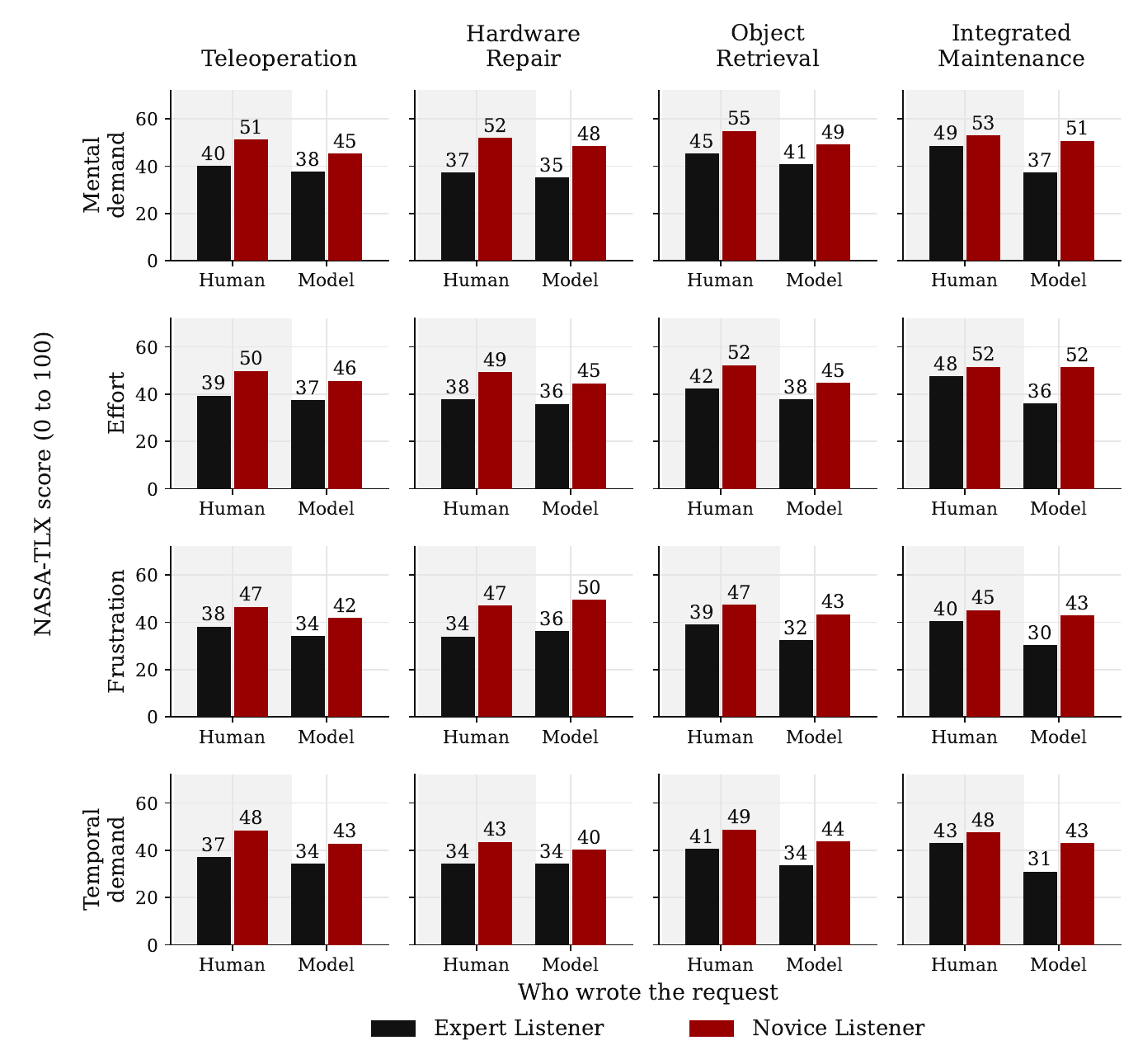}
\caption{\textbf{Four NASA-TLX subscales for human listeners evaluating human- and model-written requests.} Panels show mental demand, effort, frustration, and temporal demand across four tasks. Higher values indicate greater reported demand, effort, or frustration. Black bars represent expert listeners and red bars novice listeners. Model pools the three LLM speakers. Shading identifies the earlier human-speaker reference. Values are descriptive means from separately collected studies and do not establish between-study significance.}
\Description{Sixteen bar charts compare four workload subscales across four tasks. Each panel compares expert and novice human listeners under human-written and pooled model-written requests. Novice means exceed expert means in all displayed panels.}
\label{fig:tlx-by-task}
\end{figure*}

\paragraph{Human workload.}
Figure~\ref{fig:tlx-by-task} shows mental demand, effort, frustration, and temporal demand by task and request source. All ratings come from human listeners. Novice means exceed expert means in every displayed task--subscale panel for both request sources. Between-study differences are descriptive and do not establish that LLM requests reduce workload or cause frustration.

We administered all six NASA-TLX subscales. Four are displayed here for a more compact presentation. Perceived performance is subjective and does not determine objective task success. In the follow-up, 556 of 560 trials have questionnaires, and 206 have all six items at their initial midpoint.

\subsection{Reference-Policy Outcomes (RQ4)}

\begin{figure}[!htbp]
  \centering
  \includegraphics[width=\linewidth]{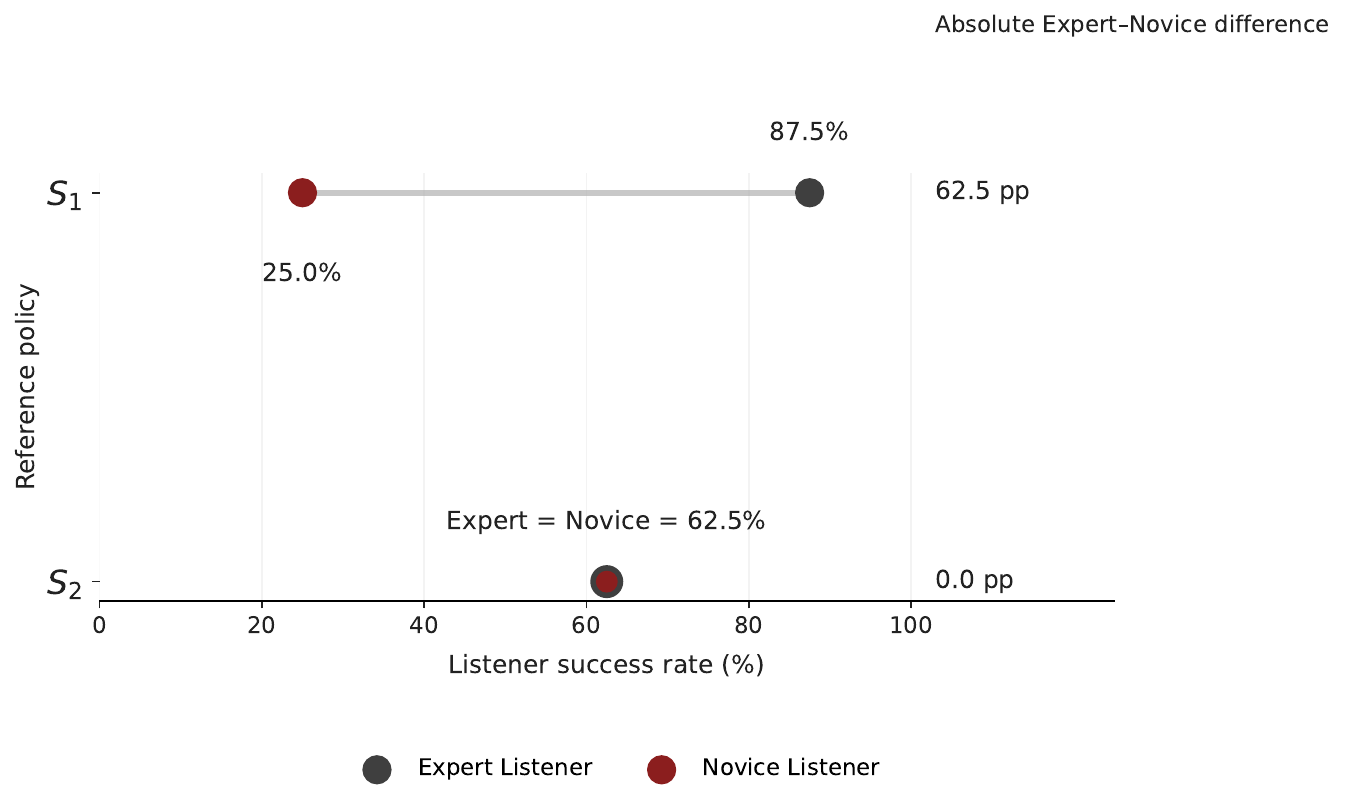}
  \caption{\textbf{Human listener success for requests selected by the reference policies.} Each policy selects one request per game scene from 321 candidates across eight scenes. Rates average the recorded success for each selected request and listener condition across scenes. $S_1$ achieves 87.5\% expert and 25.0\% novice success. $S_2$ achieves 62.5\% in each condition.}
  \Description{Expert and novice success markers for two reference policies. S1 achieves 87.5 percent expert success and 25 percent novice success. S2 achieves 62.5 percent in each condition.}
  \label{fig:policy-ladder}
\end{figure}

Figure~\ref{fig:policy-ladder} compares two corpus-based adaptations of the inverse-semantics reference policies~\cite{tellex_asking_2014}. Both select existing human-written requests. Our $S_1$ adaptation scores task information without using listener outcomes. Our $S_2$ adaptation uses a general model of listener success without separately representing expert and novice information conditions.

The candidate set contains 321 human-written requests across eight LD-HRI game scenes. Each scene is a task layout with a specified goal and reference information. Each of the four tasks has two layouts. Each policy selects one request per scene. We average the recorded outcomes for that request within each listener condition, then average across the eight scenes. Each scene therefore contributes equally to the reported rate.

For the selected requests, $S_1$ achieves 87.5\% expert success and 25.0\% novice success. $S_2$ achieves 62.5\% in each condition. Novice success is higher under $S_2$, while expert success is lower.

\begin{figure}[!htbp]
  \centering
  \includegraphics[width=0.9\linewidth]{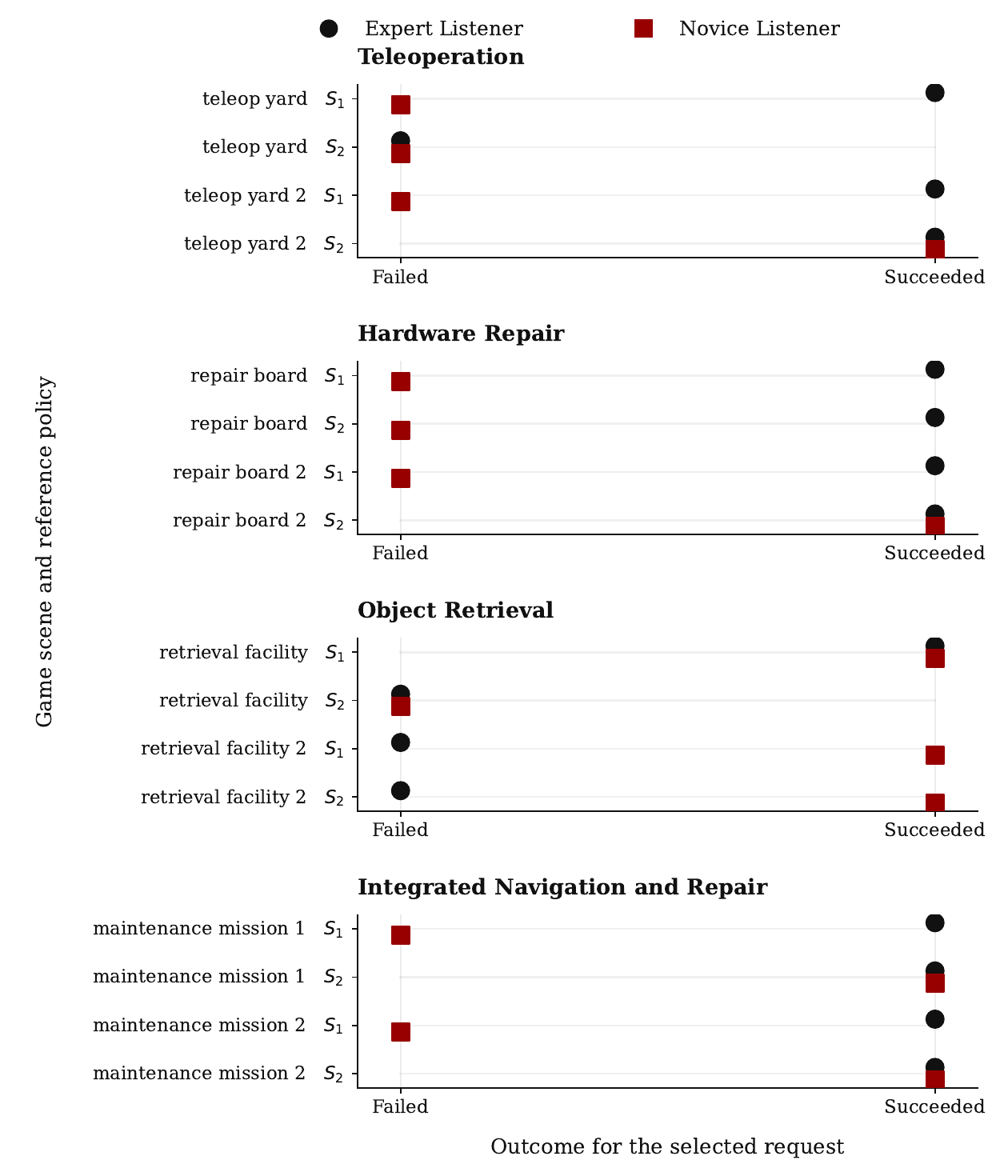}
  \caption{\textbf{Selected-request outcomes across game scenes.} Each task has two layouts. Markers show the recorded success for each selected request and listener condition. Under $S_2$, experts and novices each succeed in five scenes, but not the same five.}
  \Description{Sixteen rows grouped by task compare expert and novice outcomes for each policy and scene. Under S2, both conditions succeed in four scenes and fail in two. Only experts succeed in the first repair layout, and only novices succeed in the second retrieval layout.}
  \label{fig:policy-by-scene}
\end{figure}

The equal overall rates under $S_2$ conceal different outcomes across scenes (Figure~\ref{fig:policy-by-scene}). Both conditions succeed in four scenes and fail in two. Only experts succeed in the first Hardware Repair layout, and only novices succeed in the second Object Retrieval layout. Each condition succeeds in five scenes, but not the same five.

The policy comparison also differs across tasks. In both Integrated Navigation and Repair layouts, both conditions succeed under $S_2$, while only experts succeed under $S_1$. In the first Object Retrieval layout, both conditions succeed under $S_1$ and fail under $S_2$.

LD-HRI exposes these differences through separate results for each listener condition and game scene. Equal overall rates can conceal different failures, and a smaller expert--novice gap can accompany lower expert success. These results describe eight selections per policy and do not establish a statistically significant advantage for either policy. The fixed $S_0$ request, ``Help me,'' was not presented to human listeners and has no observed success rate in this evaluation.

\section{Discussion}

LD-HRI evaluates speakers through human listener performance. The human and LLM requests show differences in task success across listener information conditions. The properties/features of the speaker request and the information available during interaction can trigger these differences. These findings motivate the design of better help-request methods and an evaluation that considers task success, workload, and the information a person needs to act.

\subsection{Recoverability of  missing information can reduce the expert-novice success difference}

A listener's ability to recover missing information depends on the request, interface, and environment. In Hardware Repair, an unfamiliar part name cannot identify an unlabeled component. Requests that described visible properties were associated with higher novice success than requests using names alone (Figure~\ref{fig:referring-strategy}). In Object Retrieval, entering rooms reveals some scene information, although exploration may still leave the target object unclear. Integrated Maintenance combines navigation and component identification, so finding the location does not necessarily resolve the repair action. These observations suggest that help requests should prioritize information the listener cannot easily obtain during the task. Prior inverse-semantics work selects language through a model of human interpretation~\cite{tellex_asking_2014}. Our findings motivate extending this approach to distinguish information already available to a person, information available through interaction, and information the request must supply. An unfamiliar component name may need a visible description such as color and shape, while navigation instructions may need to explain the destination or the route. These are design implications from observation as tasks are inherently different.

\paragraph{Applying the task-design principles to physical robots.}
The game provides design principles for studying help requests on physical robots. Embodiment determines the assistance a robot needs from a person. For example, a mobile robot without a manipulator may need someone to reconnect a component or retrieve equipment. Recoverability determines the information that person can obtain while attempting the task. A component label may be unavailable, while an object's location may be discovered through exploration. Physical studies can vary access to component information, scene information, and control mappings, and record the information participants recover during interaction. These distinctions help separate difficulties caused by the request from difficulties caused by the physical task.

\paragraph{Implications for request-selection algorithms.}
Figure~\ref{fig:policy-ladder} illustrates the importance of reporting success in both listener conditions. The selected requests under $S_2$ produce a smaller expert--novice gap than those under $S_1$, but expert success also decreases. Gap reduction alone is therefore insufficient as an evaluation objective. Future algorithms could account for the information a listener already has, the information available through exploration, and the information the request must supply. Physical studies could cross listener information with request detail to test the benefit of explicit descriptions for novices and the communication burden for experts. Success, completion time, actions, and workload would provide complementary measures of that trade-off.

\subsection{Better Speaker Methods need to be designed for  Human and LLM Requests}

Novice success was descriptively higher with LLM-written requests across all four tasks, yet the evaluated LLM requests left a 16.4 percentage-point expert--novice gap (Figure~\ref{fig:outcome-by-task}). Thus, the model requests support many successful trials but do not remove the difficulties associated with missing reference information. Human speakers were also informed about information their listeners might lack, and their requests still left a gaps showing limitation of the requests generated, validating the importance of the HD-HRI benchmark for future research. Models can be given explicit descriptions of the intended listener's knowledge and evaluated on the requests they generate. Human studies can test instruction, examples, or feedback that help speakers explain unfamiliar terms and task information. Prior work models differences in listener knowledge computationally~\cite{takmaz_speaking_2023}. Future studies could generate separate requests for expert and novice listeners and evaluate them with LD-HRI. In our study, both listener conditions received the same request from each model for each scene.


\subsection{Workload provide an HRI metric for evaluating the impact of Success on humans}

Beyond task success evaluation by goal completion, NASA-TLX describes the demands reported by the human completing the task~\cite{hart1988development}. Novice means are higher in each of the four subscales displayed in Figure~\ref{fig:tlx-by-task} for both speaker request sources. This trend supports retaining workload alongside success in the benchmark. Small differences between the separately collected speaker studies do not establish that one request source reduces workload or causes frustration.

The full NASA-TLX measure includes physical demand and perceived performance. Shorter time spent per trial or fewer actions across all trials can reflect early failure. Each measure describes a different part of the interaction.

\subsection{Implications Beyond Robot Help Requests}

HCI research on explanations emphasizes the questions and knowledge of human receiving instructions~\cite{liao2020questioning,schaffer2019expertise}. LD-HRI applies a related concern to instructions people execute to help robots recover from failure. In healthcare, education, or search and rescue, an automated system may use terms, procedures, or locations familiar to some recipients but unfamiliar to others. The evaluation approach could be adapted to these differences by specifying the information available to each group and measuring human outcomes as an opportunity for future work.

\subsection{Limitations}

Expert and novice denote assigned access to visual information for three forms of human differences. Real helpers may differ in other forms such as experience, vocabulary, physical ability, and expectations. The browser-based tasks and one-shot textual requests do not capture speech, gesture, or opportunities for clarification during physical robot assistance. A change in interface might volunteer missing information reducing the expert-novice gap.

The human- and LLM-speaker studies used different request sets and were collected separately. Their comparison is descriptive. The LLM follow-up contains one frozen request per model and scene, with 11 or 12 participants in each model--task--condition cell. Its results characterize these requests and do not measure variability across repeated generations. Prior study exposure and changes to feedback or submission behavior during collection may affect outcomes.  Of 556 available follow-up questionnaires, 206 have all six NASA-TLX items at their initial midpoint.

Existing work uses vision-language models (VLMs) to
support robot failure recovery and describe robot
experiences~\cite{chen2024automating,wang2025ronar}.
LD-HRI provides a setting for evaluating help requests
produced by these approaches with expert and novice
human listeners. Future work could examine the impact of 
descriptions (object appearance and location) on
novice success and reduce the communication gap.
Our findings motivate the evaluation of visually grounded requests on communication gaps.

\section{Conclusion}
We introduced Listener Differences in Human-Robot Interaction (LD-HRI), a game, dataset, and benchmark for evaluating speakers through human listener performance. Across four robot-help tasks, LD-HRI captures differences in task success, workload, and interaction under expert and novice information conditions. Human- and LLM-written requests both leave substantial communication gaps. Novice success was higher with LLM-written requests across all four tasks with an expert--novice gap of 16.4\%. The reference-policy evaluation further shows that a smaller gap can accompany lower expert success, emphasizing the need to examine both listener conditions. LD-HRI provides a foundation for developing and evaluating robot help requests that account for people's task knowledge and the information they can recover from their environment. Future work can extend this evaluation to physical robots and visually grounded requests, assessing improvements in communication through human performance and experience.

\clearpage
\bibliographystyle{ACM-Reference-Format}
\bibliography{arxiv_references}

\end{document}